\documentclass[letterpaper]{article} 
\usepackage[draft]{aaai25}  
\usepackage{times}  
\usepackage{helvet}  
\usepackage{courier}  
\usepackage[hyphens]{url}  
\usepackage{graphicx} 
\usepackage{natbib}  
\usepackage{caption} 
\usepackage{algorithm}
\usepackage{algorithmic}

\usepackage{newfloat}
\usepackage{listings}
\usepackage{subcaption}
\DeclareCaptionStyle{ruled}{labelfont=normalfont,labelsep=colon,strut=off} 
\floatstyle{ruled}
\newfloat{listing}{tb}{lst}{}
\floatname{listing}{Listing}
\title{MuDoC: An Interactive Multimodal Document-grounded \\ Conversational AI System}
\author{
    Karan Taneja 
    and
    Ashok K. Goel
}
\affiliations{
    School of Interactive Computing\\
    Georgia Institute of Technology, Atlanta, GA 30332 USA\\
    \{karan.taneja,ashok.goel\}@ cc.gatech.edu
}

\usepackage{bibentry}

\begin{document}

\maketitle

\begin{abstract}
Multimodal AI is an important step towards building effective tools to leverage multiple modalities in human-AI communication.
Building a multimodal document-grounded AI system to interact with long documents remains a challenge. 
Our work aims to fill the research gap of directly leveraging grounded visuals from documents alongside textual content in documents for response generation. 
We present an interactive conversational AI agent `MuDoC' based on GPT-4o to generate document-grounded responses with interleaved text and figures.
MuDoC's intelligent textbook interface promotes trustworthiness and enables verification of system responses by allowing instant navigation to source text and figures in the documents.  
We also discuss qualitative observations based on MuDoC responses highlighting its strengths and limitations. 

\end{abstract}

\section{Introduction}

Conversational AI systems have become a part of our daily lives
and there is an increasing interest in creating multimodal dialog systems \cite{saha_towards_2018,feng_mmdialog_2023,kong_tiger_2024} to go beyond text-only conversations.
Many commercial AI systems use text-to-image generation for image outputs. 
For instance, ChatGPT \cite{openai_introducing_2022} uses DALL-E \cite{betker_improving_2023} and Google Gemini \cite{thoppilan_lamda_2022} uses Imagen \cite{saharia_photorealistic_2022}.  
These AI services can also search the web for relevant images and provide search results in the chat. 
Some recent work has also focused on interleaved text and image generation for multimodal content similar to news or blog articles \cite{dong_dreamllm_2024,zheng_minigpt-5_2024}.
In document-grounded dialog systems, AI systems have focused primarily on text-only responses based on the text in the source documents \cite{braunschweiler_evaluating_2023,taneja_jill_2024,kakar_jill_2024,olson_textbook_2025}.

For reading and comprehending long documents like textbooks, conversational AI can be a powerful tool but several key challenges exist in building such a system. 
First, the AI system must not only accommodate the past conversation in context, but also ensure that responses are sourced from documents without hallucinations.
Second, text-only responses cannot leverage visual information to convey complex ideas with simplicity of images. 
For example, imagine a user is curious about mass-pulley systems in classical mechanics. In such cases, a diagram is far more effective compared to textual descriptions.
A document-grounded AI system must go beyond text and utilize graphs, figures, and diagrams for textual explanations as well as visual content to generate a multimodal response.
Third, a trustworthy AI system should allow users to conveniently refer to the source and verify the validity of its responses.

Towards this end, we built \textbf{MuDoC}, an interactive MUltimodal DOcument-grounded Conversational AI system that addresses these challenges using publicly available resources including GPT-4o models and without any model fine-tuning.
This paper makes two main contributions: 
(i) We introduce MuDoC: a multimodal AI system to generate responses with document-grounded interleaved text and images in conversations.
(ii) We introduce MuDoC's interactive interface which enables verification of its responses through seamless navigation to source images and text in the documents to promote trustworthiness.
The subsequent sections discuss related work, MuDoC's pipeline and user interface, and some preliminary observations highlighting its strengths and limitations.

\section{Related Work}

\begin{figure*}[t]
    \centering
    \includegraphics[width=\linewidth]{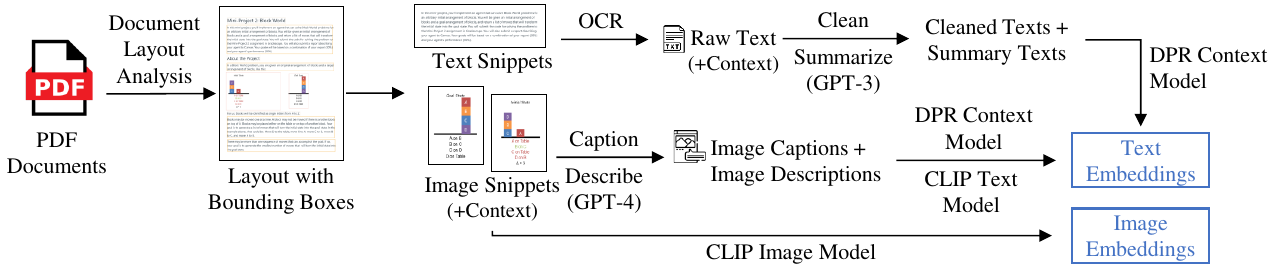}
    \caption{\textbf{Document Preprocessing:} PDF document layouts are detected to extract text and image snippets which are processed using OCR, GPT-3/4 and embedding models to create text and image embeddings for retrieval during response generation.}
    \label{fig:preprocessing}
\end{figure*}

Several multimodal dialog datasets have been proposed in the past few years 
including task-oriented datasets such as 
MMD \cite{saha_towards_2018}, 
and open-domain datasets such as 
PhotoChat \cite{zang_photochat_2021}, 
and
MMDD \cite{lee_constructing_2021}, 
MMDialog \cite{feng_mmdialog_2023}, but these datasets contain short utterances paired with a single or a few relevant images.
Several models have also been proposed for interleaved image and text generation such as
DreamLLM \cite{dong_dreamllm_2024}, 
ANOLE \cite{chern_anole_2024},
and
MiniGPT-5 \cite{zheng_minigpt-5_2024}. 
These models have applications such as visual storytelling and recipe generation.
Unlike these prior works, we wish to generate document-grounded responses in a conversational context. 
Further, generated text can rephrase original text but image generation cannot reliably create figures, diagrams, or graphs required to answer technical problems. 
Therefore, MuDoC uses text and image retrieval, and directly includes image snippets from documents for visual content in its interleaved text and image responses.
In other recent work, models for interleaved document retrieval \cite{lee_unified_2024} are being explored for similar applications.
Datasets like Doc2Dial \cite{feng_doc2dial_2020} allowed document-grounded dialog with text-only inputs and outputs.  
Previous work such as \cite{lv_kosmos-25_2024} has also explored visually-grounded chat.  

\section{MuDoC Architecture and User Interface}

Given a query and a chat context, MuDoC, our multimodal document-grounded conversational AI system, retrieves text and images from documents and answers with a multimodal response. 
Its responses contain images interleaved within text in a format similar to a typical news or a blog article. 
To create such responses, the system preprocesses documents by automatically extracting text and figures and storing their embedding representations for retrieval. 
In deployment, the system retrieves content relevant to the query using embedding-based retrieval and prompts GPT-4o to generate a response based on retrieved texts and images.
While rendering the response, figure references in the GPT-4o response are substituted with actual figures.  
MuDoC is accessed through a user interface (UI) that allows navigating to source text and figures in the documents by clicking a paragraph or a figure in its multimodal response.
In the following subsections, we discuss document preprocessing, response generation, and the UI design in detail.

\subsection{Document Preprocessing}

The preprocessing step involves building an index of text and images that can be retrieved for response generation. 
MuDoC preprocesses PDF documents and extracts texts and figures for multimodal response generation.
The resulting index contains text and image embeddings where each text embedding represents a text chunk (2000 characters long) and each image embedding represents a figure. 
Figure \ref{fig:preprocessing} outlines the preprocessing pipeline.

\textbf{Document Layout Analysis (DLA):} 
DLA is the process of identifying and classifying regions of interests in a document.
Using Region-based Convolutional Neural Networks \cite{he_mask_2017}, bounding boxes of different document components are predicted and classified into title, text, figure, list or table. 
Mask R-CNN model 
trained on the PubLayNet dataset \cite{zhong_publaynet_2019} is used to detect layout for each page. 
After extracting layout snippets in image format, each snippet (except of type `figure') is passed through the Tesseract OCR Engine
to extract raw text.
This process results in a list of document components with their page numbers, bounding boxes, component class, snippet image, and raw text.  
At this point, text components are concatenated into 2000 characters long \textit{text chunks} with at least 500 characters overlap for redundancy.

\textbf{Text Cleaning and Summarization:}
The raw text from the OCR model can have typographical errors and poor formatting.
For text snippets, these errors are corrected by prompting GPT-3.5 
(gpt-3.5-turbo-16k-0613)
to generate error-free neatly-formatted \textit{cleaned text}. 
Additionally, GPT-3.5 is prompted to generate \textit{summary text}, which provides a compact version of the text for additional text embeddings, and is only generated if there are more than 1000 characters in the OCR raw text.
All text from previous, current, and next page of first snippet in the text chunk is also provided as context in this process 
which simplifies correction of any ambiguous typographical errors or complex formatting.

\textbf{Image Captions and Descriptions:} 
For image snippets, image captions and descriptions are generated for later use as an input in the response generation along with images. 
Image captions and descriptions provide more context to the retrieval results and support referring to images in response text.
To generate these, the raw image snippet to GPT-4 (gpt-4-vision-preview) is provided along with all text from previous, current and next page as context. 

\textbf{DPR and CLIP Embeddings:} 
Dense Passage Retrieval (DPR) \cite{karpukhin_dense_2020} consists of a query encoder and a context encoder which are aligned to give similar embeddings if a context can answer a query. 
DPR context encoder is used to pre-compute context embeddings for raw, cleaned, and summary text, and image captions and descriptions.
These embeddings are stored for the retrieval process described later.
Contrastive Language-Image Pre-training (CLIP) \cite{radford_learning_2021} models, similar to DPR models, have two aligned models where one model encodes images and the other encodes texts. 
CLIP image model
is used to pre-compute image embeddings as well as text embeddings for image captions and descriptions. 

\subsection{Response Generation} 

\begin{figure*}[t]
    \centering
    \includegraphics[width=\linewidth]{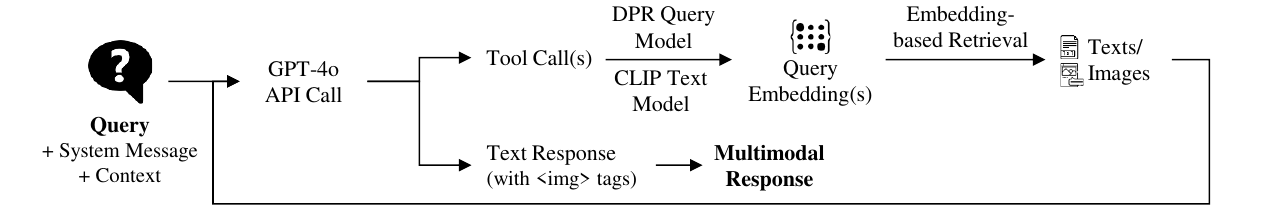}
    \caption{\textbf{Response Generation:} GPT-4o tool calls feature is used to create text and image retrieval queries. Outputs from embedding-based retrieval are used for response generation and image references are replaced with actual images.}
    \label{fig:response-generation}
\end{figure*}

\textbf{Content Retrieval:}
Upon receiving a query, the system may perform text and/or image retrieval before generating a response.
For text retrieval, the query embedding is computed using the DPR encoder model.
Next, cosine similarities between the query DPR embedding and pre-computed context embeddings are calculated for all text chunks. 
Maximum cosine similarity across raw, cleaned, and summary text is used as the text chunk score, and the top-5 chunks are used as retrieval output. 
For image retrieval, both DPR and CLIP embeddings are used to determine the top-5 images. 
All cosine similarities between query (CLIP/DPR) embeddings and images (CLIP), captions (CLIP/DPR) and descriptions (CLIP/DPR) embeddings are computed. Maximum CLIP and DPR cosine similarities across all representations are determined, and the mean of CLIP and DPR scores is used as the final score for each image.   

\textbf{GPT-4o Tool Calls and Prompting:} 
MuDoC uses GPT-4o (gpt-4o-2024-08-06) to generate responses and its \textit{tool calls} feature with structured outputs
to query the retrieval functions for text and images (see Figure \ref{fig:response-generation}). 
For MuDoC, GPT-4o API is set up to return a text response or a tool call request with a search query for text or image retrieval. 
In the former case, the response is directly forwarded to the user. In the latter case, retrieval is performed from on our end and the results are returned with another API call.
The \textit{system message} contains instructions to (i) answer only using texts and images retrieved using search tools, (ii) allow multiple search queries before responding, (iii) use visuals but only if they are appropriate and useful, (iv) refer to image using HTML image and caption tag and a description in the text, and (v) use a style similar to blogs or academic texts.
The context contains past user messages, system responses, tool calls, and retrieval results, together up to most recent 64K characters in order to limit the input size and API costs. 
GPT-4o API does not allow images in tool call outputs, so the images are appended as a user message immediately after the image retrieval output with image filenames, captions and descriptions.
The system message described earlier ensures that API response contains HTML image tags with filenames in the retrieval output. 
While rendering the response, these image tags are replaced with actual images and add generated captions below each image.

\subsection{User Interface}

\begin{figure*}[t]
    \centering
    \begin{subfigure}[b]{0.71\textwidth}
        \centering
        \includegraphics[width=\textwidth]{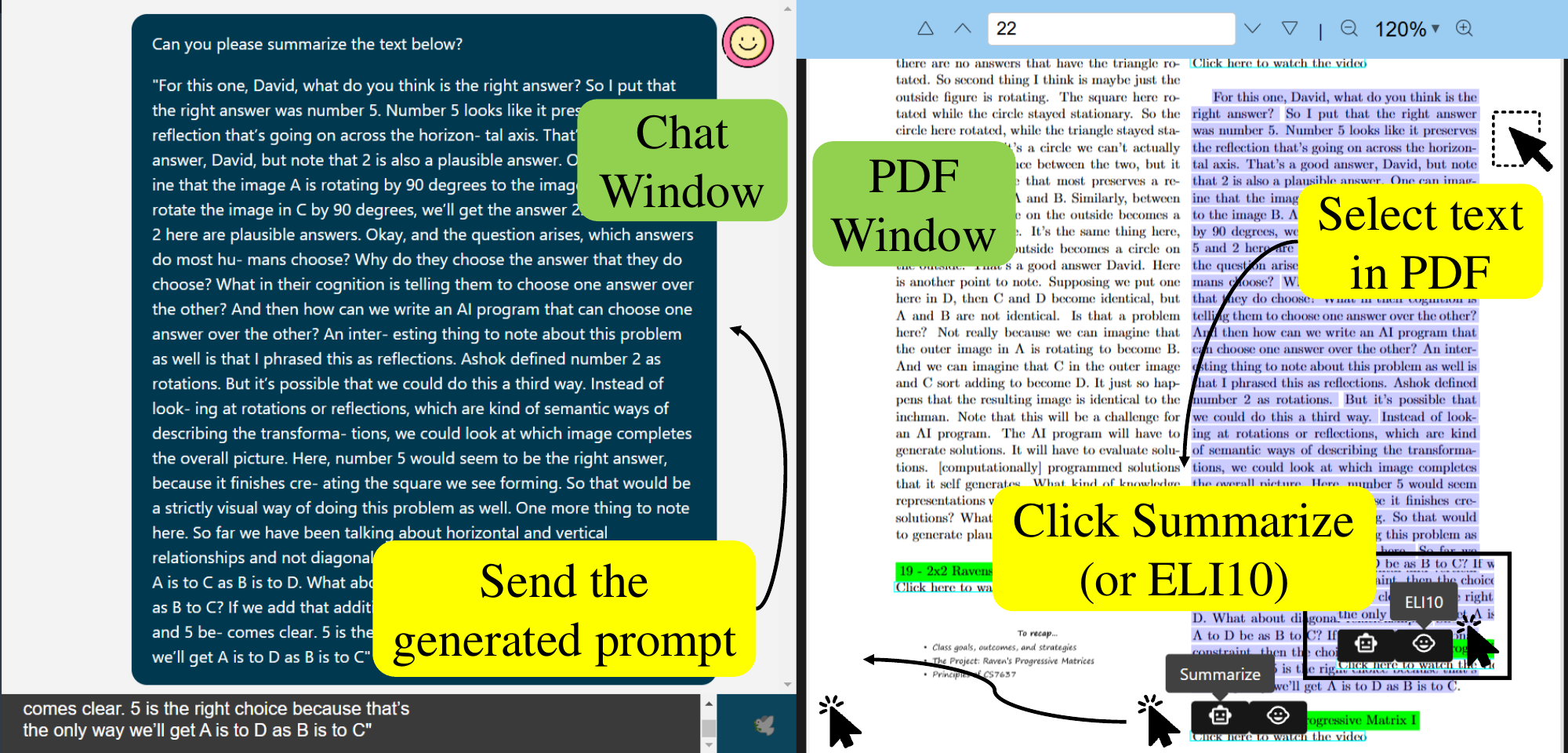} 
        \caption{Chat-PDF Display and Summarize/ELI10}
        \label{fig:ui-summarize}
    \end{subfigure}
    \hfill 
    \begin{subfigure}[b]{0.49\textwidth}
        \centering
        \includegraphics[width=\textwidth]{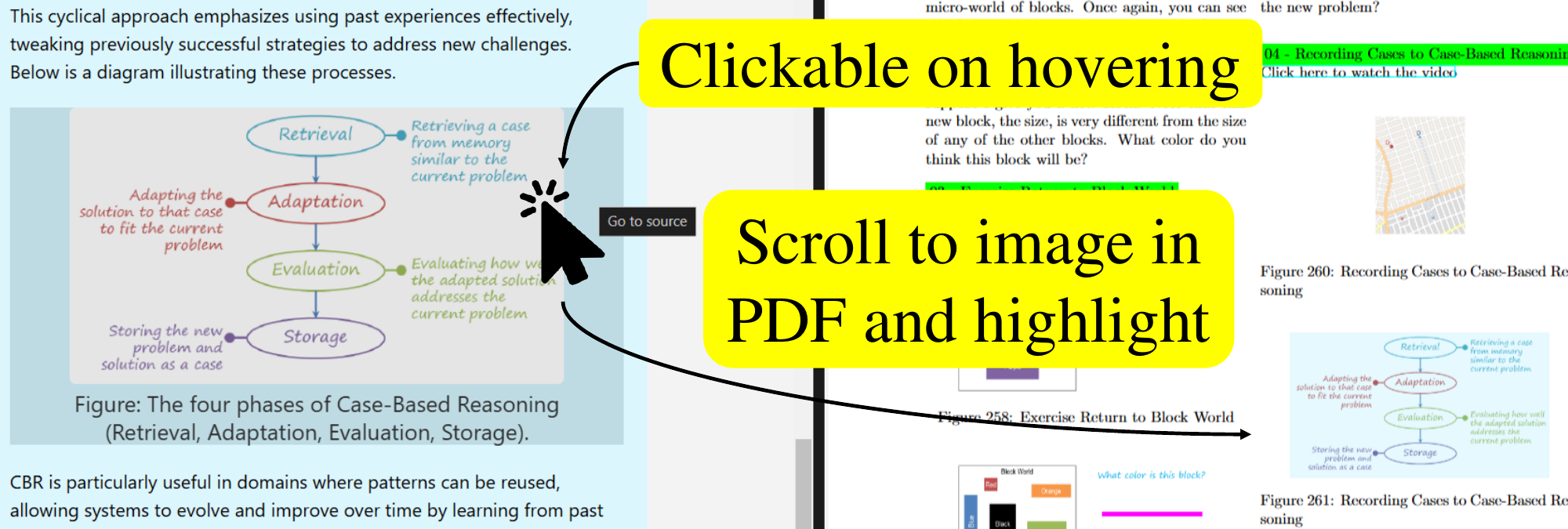} 
        \caption{Clickable Images for Navigation}
        \label{fig:ui-image-highlight}
    \end{subfigure}
    \hfill
    \begin{subfigure}[b]{0.49\textwidth}
        \centering
        \includegraphics[width=\textwidth]{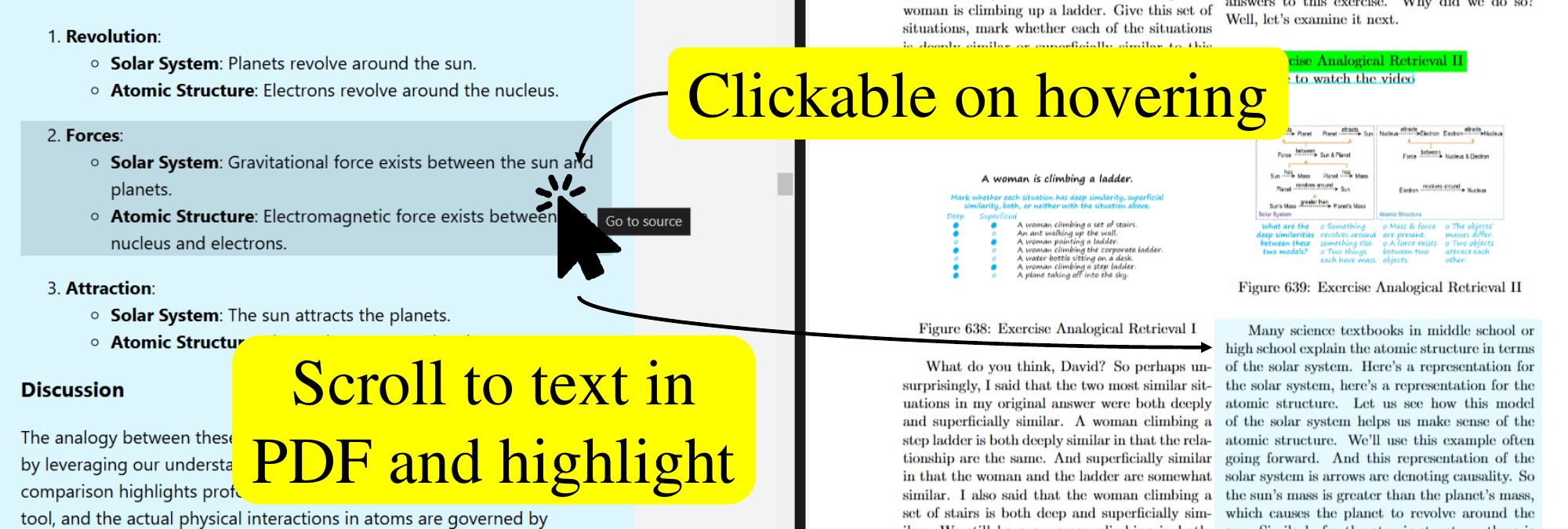}
        \caption{Clickable Text for Navigation}
        \label{fig:ui-text-highlight}
    \end{subfigure}
    \caption{\textbf{UI Features:} Chat-PDF display in (a) shows chat area on left, text box on bottom-left, and PDF on right. Yellow boxes and arrows describe features including summarize, Explain-it-Like-I'm-10 (ELI10), PDF navigation using text and images.}
    \label{fig:ui-interface}
\end{figure*}

MuDoC's UI provides an intelligent textbook experience through some powerful features that allow users to interactively explore documents (see Figure \ref{fig:ui-interface}). 

\textbf{UI Implementation:}
The UI is implemented using the ReactJS framework and the entire application is served through a Flask server.
Token streaming is enabled from OpenAI API through our server to the front end using HTTP Server-Sent Events (SSE) to reduce the time to first token.
During retrieval and API calls, chat messages indicating the system status such as `Gathering information', `Retrieving text/images for [query]', and `Generating a response' are displayed. 
The PDF display features are implemented using PDF.js
PDF-viewer supported by Mozilla.

\textbf{Summarize and Explain-it-Like-I'm-10 (ELI10):}
As shown in Figure \ref{fig:ui-summarize}, users see a chat window on the left side of the display with a text-box to write queries at the bottom,
and a PDF document on the right.
To summarize a piece of text, a user can drag their cursor to select a paragraph and see the two options viz. `Summarize' and `ELI10'.
Upon selecting an option, a prompt is created and pushed to the chat text-box where it can be edited and sent to the back-end.
The summarize prompt leads to a shorter text which helps in quick reading while the ELI10 feature leads to a simpler answer with little jargon which supports readability of the content.
These features can support content exploration in an interactive method and make it engaging for users to read long and complex texts.

\textbf{Navigation Using Images and Texts:} 
When a user hovers their cursor over an image or a paragraph in a multimodal response, the image appears to be clickable through brightness change and tooltip text as shown in Figure \ref{fig:ui-image-highlight} and \ref{fig:ui-text-highlight}. 
Upon clicking an image or a paragraph, the PDF scrolls to the relevant page where the figure appears or a similar text is found and the figure or the text is highlighted for three seconds.
The bounding box information of snippets stored during the preprocessing step is used for images. 
But, unlike images, paragraphs are typically rephrased and not directly borrowed from the document.
So, a post-processing step is performed after response generation where paragraphs in the response are mapped to raw text snippets based on cosine similarity between their DPR context embeddings. 
Paragraphs with a length below 100 characters or similarity scores below a threshold are not mapped and these paragraphs cannot be clicked for navigation.
Navigation using images and paragraphs allow users to understand the context in which a figure or a text shows up in the document. 
Users can also choose to read surrounding content to delve deeper by reading from the document which is a more complete and trustworthy source unlike AI-generated text which can be hallucinated or lack context.

\section{Preliminary Observations}

For MuDoC's initial testing, we used Knowledge-based AI ebook \cite{goel_knowledge-based_2018} with 357 pages and numerous figures illustrating AI concepts and algorithms.
MuDoC responses contain long texts and multiple figures which leads to long conversations that cannot be included here.
We provide some illustrative examples of conversations with MuDoC at
\texttt{tinyurl.com/MuDoCSamples}.
We asked some subjects to use MuDoC to learn AI concepts from the ebook to test out the system and get their feedback. 

Based on these conversations, we observed that MuDoC responses contain relevant images from the ebook interleaved within text. 
MuDoC retrieves images that address the query and GPT-4o supports refined selection from top-5 retrieved images.  
A user pointed out that they gravitated towards visuals over text in MuDoC responses.
Navigation by clicking images or paragraphs and instantly seeing relevant content in the 357-page ebook was highly appreciated. 
Summarize and ELI10 features using text selection also supported reading by simplifying the prompting process.
From the qualitative feedback, we found that the figure placement in responses is not always optimal. 
In many cases, visualizations were presented in a separate section with additional text which lead to extremely long responses. 
While most figures were helpful, some figures were only loosely connected to the topics under discussion. 
Such extra information can be distracting for users and can reduce the system's effectiveness as a productivity tool.  
We also observed that a text describing an image was hallucinated and, upon further investigation, we found that these hallucinations occur more often when multiple images are provided to GPT-4o as a part of a single query.
This suggests that reducing the number of retrieved images maybe improve visual coherence in MuDoC's responses.
Finally, paragraphs is some responses were mapped to short headings in the document instead of a complete text snippet which can lead to perceptual difficulty in navigation.
Overall, despite its current limitations, initial observations are encouraging and indicate that MuDoC is a promising step towards making multimodal conversational AI accessible and trustworthy for widespread, everyday use with documents like long textbooks.

\section{Conclusions and Future Work}
We presented MuDoC, an interactive multimodal document-grounded AI system, with interleaved text and image responses. 
MuDoC leverages GPT-4o for document-grounding and multimodality, and its user interface allows quick navigation with clickable figures and paragraphs which increases trust and verifiability of responses.
Despite its current limitations, it demonstrates that multimodal dialog can be a powerful interaction medium for long documents and seems promising as an effective productivity tool. 
In future work, we will quantitatively evaluate its performance in terms of response quality and its effectiveness in education as a tool for learners.  

\noindent\textbf{Acknowledgements:}
This work is supported by a grant from NSF (Award \#2247790) as well as a grant from Bill and Melinda Gates Foundation.
We also thank Audrey Olson and Anjali Singh for their support and feedback.

\bibliography{main}

\begin{thebibliography}{24}
\providecommand{\natexlab}[1]{#1}

\bibitem[{Betker et~al.(2023)Betker, Goh, Jing, Brooks, Wang, Li, Ouyang, Zhuang, Lee, Guo, Manassra, Dhariwal, Chu, Jiao, and Ramesh}]{betker_improving_2023}
Betker, J.; Goh, G.; Jing, L.; Brooks, T.; Wang, J.; Li, L.; Ouyang, L.; Zhuang, J.; Lee, J.; Guo, Y.; Manassra, W.; Dhariwal, P.; Chu, C.; Jiao, Y.; and Ramesh, A. 2023.
\newblock Improving {Image} {Generation} with {Better} {Captions}.

\bibitem[{Braunschweiler et~al.(2023)Braunschweiler, Doddipatla, Keizer, and Stoyanchev}]{braunschweiler_evaluating_2023}
Braunschweiler, N.; Doddipatla, R.; Keizer, S.; and Stoyanchev, S. 2023.
\newblock Evaluating {Large} {Language} {Models} for {Document}-grounded {Response} {Generation} in {Information}-{Seeking} {Dialogues}.
\newblock In \emph{{Association for Computational Linguistics} {Workshop} on {Taming} {Large} {Language} {Models}: {Controllability} in the era of {Interactive} {Assistants}}, 46--55.

\bibitem[{Chern et~al.(2024)Chern, Su, Ma, and Liu}]{chern_anole_2024}
Chern, E.; Su, J.; Ma, Y.; and Liu, P. 2024.
\newblock {ANOLE}: {An} {Open}, {Autoregressive}, {Native} {Large} {Multimodal} {Models} for {Interleaved} {Image}-{Text} {Generation}.
\newblock ArXiv:2407.06135 [cs].

\bibitem[{Dong et~al.(2024)Dong, Han, Peng, Qi, Ge, Yang, Zhao, Sun, Zhou, Wei, Kong, Zhang, Ma, and Yi}]{dong_dreamllm_2024}
Dong, R.; Han, C.; Peng, Y.; Qi, Z.; Ge, Z.; Yang, J.; Zhao, L.; Sun, J.; Zhou, H.; Wei, H.; Kong, X.; Zhang, X.; Ma, K.; and Yi, L. 2024.
\newblock {DreamLLM}: {Synergistic} {Multimodal} {Comprehension} and {Creation}.
\newblock In \emph{{ International Conference on Learning Representations} 2024 {Spotlight}}.

\bibitem[{Feng et~al.(2023)Feng, Sun, Xu, and et~al.}]{feng_mmdialog_2023}
Feng, J.; Sun, Q.; Xu, C.; and et~al. 2023.
\newblock {MMDialog}: {A} {Large}-scale {Multi}-turn {Dialogue} {Dataset} {Towards} {Multi}-modal {Open}-domain {Conversation}.
\newblock In \emph{{Association for Computational Linguistics} 2023}, 7348--7363.

\bibitem[{Feng et~al.(2020)Feng, Wan, Gunasekara, Patel, Joshi, and Lastras}]{feng_doc2dial_2020}
Feng, S.; Wan, H.; Gunasekara, C.; Patel, S.; Joshi, S.; and Lastras, L. 2020.
\newblock doc2dial: {A} {Goal}-{Oriented} {Document}-{Grounded} {Dialogue} {Dataset}.
\newblock In \emph{{Empirical} {Methods} in {Natural} {Language} {Processing} 2020}, 8118--8128. Online.

\bibitem[{Goel and Joyner(2018)}]{goel_knowledge-based_2018}
Goel, A.; and Joyner, D. 2018.
\newblock \emph{Knowledge-based {Artificial} {Intelligence}: {Cognitive} {Systems} (e-book)}.
\newblock Georgia Institute of Technology.

\bibitem[{He et~al.(2017)He, Gkioxari, Dollár, and Girshick}]{he_mask_2017}
He, K.; Gkioxari, G.; Dollár, P.; and Girshick, R. 2017.
\newblock Mask {R}-{CNN}.
\newblock In \emph{{International Conference on Computer Vision} 2017}, 2980--2988.

\bibitem[{Kakar et~al.(2024)Kakar, Maiti, Taneja, Nandula, Nguyen, Zhao, Nandan, and Goel}]{kakar_jill_2024}
Kakar, S.; Maiti, P.; Taneja, K.; Nandula, A.; Nguyen, G.; Zhao, A.; Nandan, V.; and Goel, A. 2024.
\newblock Jill {Watson}: {Scaling} and {Deploying} an {AI} {Conversational} {Agent} in {Online} {Classrooms}.
\newblock In \emph{{ITS} 2024}, 78--90.

\bibitem[{Karpukhin et~al.(2020)Karpukhin, Oguz, Min, Lewis, Wu, Edunov, Chen, and Yih}]{karpukhin_dense_2020}
Karpukhin, V.; Oguz, B.; Min, S.; Lewis, P.; Wu, L.; Edunov, S.; Chen, D.; and Yih, W.-t. 2020.
\newblock Dense {Passage} {Retrieval} for {Open}-{Domain} {Question} {Answering}.
\newblock In \emph{{Empirical Methods in Natural Language Processing} 2020}, 6769--6781.

\bibitem[{Kong et~al.(2024)Kong, Wang, Feng, Wang, and Zhang}]{kong_tiger_2024}
Kong, F.; Wang, P.; Feng, S.; Wang, D.; and Zhang, Y. 2024.
\newblock {TIGER}: {A} {Unified} {Generative} {Model} {Framework} for {Multimodal} {Dialogue} {Response} {Generation}.
\newblock In \emph{{Joint International Conference on Computational Linguistics, Language Resources and Evaluation} 2024}, 16135--16141.

\bibitem[{Lee et~al.(2024)Lee, Ko, Baek, Jeong, and Hwang}]{lee_unified_2024}
Lee, J.; Ko, J.; Baek, J.; Jeong, S.; and Hwang, S.~J. 2024.
\newblock Unified {Multimodal} {Interleaved} {Document} {Representation} for {Retrieval}.
\newblock ArXiv:2410.02729 [cs].

\bibitem[{Lee et~al.(2021)Lee, Shin, Choo, Choi, and Myaeng}]{lee_constructing_2021}
Lee, N.; Shin, S.; Choo, J.; Choi, H.-J.; and Myaeng, S.-H. 2021.
\newblock Constructing {Multi}-{Modal} {Dialogue} {Dataset} by {Replacing} {Text} with {Semantically} {Relevant} {Images}.
\newblock In \emph{{ACL} {International Joint Conference on Natural Language Processing} 2021}, 897--906.

\bibitem[{Lv et~al.(2024)Lv, Huang, Chen, Zhao, Jia, Cui, Ma, Chang, Huang, Wang, Dong, Luo, Wu, Wang, Zhang, and Wei}]{lv_kosmos-25_2024}
Lv, T.; Huang, Y.; Chen, J.; Zhao, Y.; Jia, Y.; Cui, L.; Ma, S.; Chang, Y.; Huang, S.; Wang, W.; Dong, L.; Luo, W.; Wu, S.; Wang, G.; Zhang, C.; and Wei, F. 2024.
\newblock {KOSMOS}-2.5: {A} {Multimodal} {Literate} {Model}.
\newblock ArXiv:2309.11419 [cs].

\bibitem[{Olson, Maiti, and Goel(2025)}]{olson_textbook_2025}
Olson, A.; Maiti, P.; and Goel, A. 2025.
\newblock The {Textbook} of {Tomorrow}: {Rethinking} {Course} {Material} {Interfacing} in the {Era} of {GPT}.
\newblock ArXiv:2501.03618 [cs].

\bibitem[{{OpenAI}(2022)}]{openai_introducing_2022}
{OpenAI}. 2022.
\newblock Introducing {ChatGPT}. http://openai.com/index/chatgpt. Accessed: 2025-02-13.

\bibitem[{Radford et~al.(2021)Radford, Kim, Hallacy, Ramesh, Goh, Agarwal, Sastry, Askell, Mishkin, Clark, Krueger, and Sutskever}]{radford_learning_2021}
Radford, A.; Kim, J.~W.; Hallacy, C.; Ramesh, A.; Goh, G.; Agarwal, S.; Sastry, G.; Askell, A.; Mishkin, P.; Clark, J.; Krueger, G.; and Sutskever, I. 2021.
\newblock Learning {Transferable} {Visual} {Models} {From} {Natural} {Language} {Supervision}.
\newblock In \emph{{International Conference on Machine Learning} 2021}, 8748--8763.

\bibitem[{Saha, Khapra, and Sankaranarayanan(2018)}]{saha_towards_2018}
Saha, A.; Khapra, M.~M.; and Sankaranarayanan, K. 2018.
\newblock Towards building large scale multimodal domain-aware conversation systems.
\newblock In \emph{{Association for the Advancement of Artificial Intelligence} {Symposium} on {Educational} {Advances} in {Artificial} {Intelligence} 2018}, 696--704.

\bibitem[{Saharia et~al.(2022)Saharia, Chan, Saxena, Li, Whang, Denton, Ghasemipour, Ayan, Mahdavi, Lopes, Salimans, Ho, Fleet, and Norouzi}]{saharia_photorealistic_2022}
Saharia, C.; Chan, W.; Saxena, S.; Li, L.; Whang, J.; Denton, E.; Ghasemipour, S. K.~S.; Ayan, B.~K.; Mahdavi, S.~S.; Lopes, R.~G.; Salimans, T.; Ho, J.; Fleet, D.~J.; and Norouzi, M. 2022.
\newblock Photorealistic {Text}-to-{Image} {Diffusion} {Models} with {Deep} {Language} {Understanding}.
\newblock In \emph{{Neural Information Processing Systems} 2022}.

\bibitem[{Taneja et~al.(2024)Taneja, Maiti, Kakar, Guruprasad, Rao, and Goel}]{taneja_jill_2024}
Taneja, K.; Maiti, P.; Kakar, S.; Guruprasad, P.; Rao, S.; and Goel, A.~K. 2024.
\newblock Jill {Watson}: {A} {Virtual} {Teaching} {Assistant} {Powered} by {ChatGPT}.
\newblock In \emph{{International Conference on Artificial Intelligence in Education} 2024}, 324--337.

\bibitem[{Thoppilan et~al.(2022)Thoppilan, De~Freitas, Hall, Shazeer, Kulshreshtha, Cheng, Jin, and {others}}]{thoppilan_lamda_2022}
Thoppilan, R.; De~Freitas, D.; Hall, J.; Shazeer, N.; Kulshreshtha, A.; Cheng, H.-T.; Jin, A.; and {others}. 2022.
\newblock {LaMDA}: {Language} {Models} for {Dialog} {Applications}.
\newblock ArXiv:2201.08239 [cs].

\bibitem[{Zang et~al.(2021)Zang, Liu, Wang, Song, Zhang, and Chen}]{zang_photochat_2021}
Zang, X.; Liu, L.; Wang, M.; Song, Y.; Zhang, H.; and Chen, J. 2021.
\newblock {PhotoChat}: {A} {Human}-{Human} {Dialogue} {Dataset} {With} {Photo} {Sharing} {Behavior} {For} {Joint} {Image}-{Text} {Modeling}.
\newblock In \emph{{ACL} {International Joint Conference on Natural Language Processing} 2021}, 6142--6152.

\bibitem[{Zheng, He, and Wang(2024)}]{zheng_minigpt-5_2024}
Zheng, K.; He, X.; and Wang, X.~E. 2024.
\newblock {MiniGPT}-5: {Interleaved} {Vision}-and-{Language} {Generation} via {Generative} {Vokens}.
\newblock ArXiv:2310.02239 [cs].

\bibitem[{Zhong, Tang, and Jimeno~Yepes(2019)}]{zhong_publaynet_2019}
Zhong, X.; Tang, J.; and Jimeno~Yepes, A. 2019.
\newblock {PubLayNet}: {Largest} {Dataset} {Ever} for {Document} {Layout} {Analysis}.
\newblock \emph{International Conference on Document Analysis and Recognition 2019}, 1015--1022.

\end{thebibliography}

\end{document}